# Benchmarking Quantized Neural Networks on FPGAs with FINN

Quentin Ducasse, Pascal Cotret, Loïc Lagadec
*Lab-STICC, ENSTA Bretagne*
Brest, France
firstname.lastname@ensta-bretagne.org

Robert Stewart
*Heriot-Watt University*
Edinburgh, United Kingdom
r.stewart@hw.ac.uk

*Abstract*—The ever-growing cost of both training and inference for state-of-the-art neural networks has brought literature to look upon ways to cut off resources used with a minimal impact on accuracy. Using lower precision comes at the cost of negligible loss in accuracy. While training neural networks may require a powerful setup, deploying a network must be possible on low-power and low-resource hardware architectures. Reconfigurable architectures have proven to be more powerful and flexible than GPUs when looking at a specific application. This article aims to assess the impact of mixed-precision when applied to neural networks deployed on FPGAs. While several frameworks exist that create tools to deploy neural networks using reduced-precision, few of them assess the importance of quantization and the framework quality. *FINN* and *Brevitas*, two frameworks from Xilinx labs, are used to assess the impact of quantization on neural networks using 2 to 8 bit precisions and weights with several parallelization configurations. Equivalent accuracy can be obtained using lower-precision representation and enough training. However, the compressed network can be better parallelized allowing the deployed network throughput to be 62 times faster. The benchmark set up in this work is available in a public repository (https://github.com/QDucasse/nn_benchmark).

*Index Terms*—Machine Learning, Neural Networks, Mixed Precision, FPGA

## I. Introduction

Convolutional Neural Networks (CNNs) have gained a rising interest in the field of machine learning. Their deployment to embedded devices is now at the heart of the discussions. Many low-power low-memory IoT devices require running image classification or voice recognition tasks. Running a CNN inference requires the host device to both hold the network itself and then run the billion of operations an inference implies. Reducing and optimizing this memory and power footprint has been the focus of literature in recent years.

Using a neural network requires a previous phase of *Training* that should only be done once and can be performed on powerful architectures (clusters of GPUs for example). Then the network goes through a *Deployment* phase to the target architecture which can be a low-cost device. Moreover, the deployed network should then be able to perform an *Inference* phase each time a new instance needs to be classified. While *Training* is only done once without restriction on the running device, *Inference* is repeated through the application and should be more optimized for low-resources devices.

While architecture such as CPUs or GPUs are used to perform the different steps, FPGAs and their natural reduced-precision capabilities seem to perform the best. Mixed-precision (or reduced precision) consists in using numbers representations that require less space to map a number, often at the cost of some precision loss. There is a tight relation between the number representations used (*64-bits* double-precision, *32-bits* single-precision or *16-bits* half-precision) and the costs of the operations in terms of memory and power [1], [2]. Figure 1 shows that increasing precision on the numbers used increases the relative energy cost accordingly. An 8-bit addition costs 0.03 pJ whereas the same addition in 32-bit floating-point costs 0.9 pJ or 30 times more.

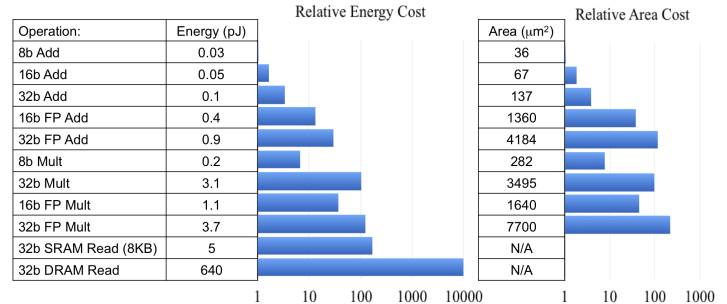

Fig. 1: Cost of operations in certain representations [1], [2].

In the field of machine learning, this results in a new type of networks named *Quantized Neural Networks (QNNs)* [3] and their extreme versions, *Binarized Neural Networks (BNNs)* [4], [5] or *XNOR-Net* [6]. This type of network is gaining momentum as the overgrowing size of network architectures needs to perform the best on the smallest end architectures. Google Tensorflow Lite (tensorflow.org/lite) can reduce full-precision networks to 8-bit counterparts. Intel Distiller works as a compression framework over *PyTorch* [7]. Moreover, recent iterations over *PyTorch* itself to provide the users with means to define QNNs [8]. Those examples highlight the need for quantization, even in industry led neural network frameworks.

*Quantization* comes as one of the many optimization methods for CNNs. It consists of a type of network compression and uses *data type* as the focus for its compression. Other compression methods might involve *pure compression* through a hash function grouping similar connections and weights [9] or *pruning*, where a trade-off objective is set and a pruning routine is used to remove the least important neurons until the objective is met [10], [11]. Those methods can keep an important accuracy while reducing effectively the network size. Even if recent efforts can be noticed in the literature with a combination of several optimization techniques [12], this paper will exclusively focus on quantization.

On the other hand, neural network frameworks are developed focusing on FPGA deployment. Early iterations can be noted [13], [14] but are not actively maintained or open-source, therefore restricting their use. However, *FPGAConvNet* [15] provides a way to deploy CNNs on FPGAs. Xilinx *FINN* [16] is another framework to port neural networks on FPGAs and has been extended to QNNs and BNNs [17], as well as Long-Short Term Memory Neural Networks (LSTM) [18]. The support of the different networks come through *Brevitas* [19], a quantization-aware trainer facility built as a drop-in replacement of *PyTorch* and tightly linked to *FINN*.

While other optimization methods exist such as pruning, this paper focuses on the trade-off between accuracy and FPGA resource requirements that the underlying number representation presupposes. The design space explored is the evaluation of QNNs with varying bit-width values for their weights and activations. This design space is important to benchmark as very few initiatives exist to compare frameworks and quantizations between them [20], [21].

In this paper, we present the development and setup of an overlay of the Xilinx framework *FINN* and the Quantized Neural Network (QNN) trainer facility *Brevitas* to port QNNs on FPGAs. This overlay is then used to assess the impact of quantization on several metrics, both from an accuracy and hardware utilization perspective. The closest related work to our benchmarking detailed in Section III is the work proposed by Bacchus et al. [22]: this paper proposes a more detailed analysis where several parallelization configurations are explored in order to find the best compromise between quantization parameters and neural networks hardware configurations. Moreover, the framework used by Bacchus et al. has now been archived by the Xilinx team while *FINN* and *Brevitas* are in active development.

The rest of the paper is organized as follows. Section II presents the background of the project, the *FINN* workflow as well as quantization methods through the trainer *Brevitas*. Section III presents the experimental setup (topology, platform, methodology) as well as the results of the benchmarking. Conclusions are given in Section IV with future works aspirations.

## II. BACKGROUND

### A. FINN Workflow

While several initiatives have been created to port QNNs on FPGAs, few benchmarks are provided between the frameworks. Moreover, few projects can be reused outside of the authors initial work due to the lack of open-source code. A need to benchmark the different FPGA frameworks emerges as there are few comparisons between actual FPGA implementations. Only *FINN*, maintained by Xilinx can be used freely and is in active development. The benchmark will be performed on this particular framework for now. *FINN* uses the model produced by *Brevitas*, a quantized neural networks trainer and designer developed by Xilinx as well on top of *PyTorch*. By first going through *Brevitas*, the quantized neural network is translated into the intermediate representation, *ONNX*. The resulting intermediate representation can then be imported by *FINN*. *FINN* will perform several transformations and deploy part of the resulting graph on the associated FPGA. To summarize, the workflow is the following: **Brevitas** for *Training*, **ONNX** as the *Intermediate Representation* and **FINN** for *Network Restructuration*, *HLS Synthesis* and *Deployment*.

*1) Brevitas:* *Brevitas* (xilinx.github.io/brevitas) has been developed with the idea of corresponding to a drop-in replacement of *PyTorch*. This means that it ensures that *PyTorch* functionalities will be preserved, even when working with reduced-precision layers. *Brevitas* implements a set of building blocks to model a reduced precision hardware data-path at training time. While partially biased towards modeling data-flow-style, very low-precision implementations, building blocks can be parameterized and assembled together to target all sorts of reduced precision hardware. Once the neural network is defined and trained, *Brevitas* provides a way to export it as the intermediate representation. This representation is the focus of the next subsection and will be looked into in more details later on. *Brevitas* uses specific annotations to define the quantized layers.

The intermediate representation, called *ONNX* does not provide a way to represent layers with precisions under 8 bits: this is why this system of custom annotations has been created. These specific annotations are then be used by *FINN* to detect the quantization and perform adequate transformations on the layers. An example of the linear layer (*FC*) and the hyperbolic tangent activation function using *PyTorch* and *Brevitas* is shown in Figure 2.

```python
import torch
import torch.nn as nn
import brevitas.nn as bnn
# Classic sequence in PyTorch
classic_sequence = nn.Sequential(
  nn.Linear(120, 84),
  nn.Tanh()
)
# Quantized sequence in Brevitas
quantized_sequence = nn.Sequential(
  bnn.QuantLinear(120, 10, bit_width = 3),
  bnn.QuantTanh(bit_width = 5)
)
```

Fig. 2: Layers comparison between PyTorch and *Brevitas*.

This piece of code presents the definition of two sequences of layers as defined in the `Sequential` class in *Pytorch*. In addition, *Brevitas* quantized layers allow the user to choose the bitwidth (i.e. precision) needed with the additional `bit_width` argument.

*2) ONNX:* The Open Neural Network Exchange *ONNX* (onnx.ai) project provides an open-source format for artificial intelligence and machine learning models. This is done by defining an extensible computation graph model, as well as definitions of operators and data types. *ONNX* is widely supported in different machine learning frameworks and tools. The main goal behind its design and development was to enable interoperability between different machine learning frameworks to streamline the path between research and production.

*3) FINN:* *FINN* workflow (xilinx.github.io/finn) itself is decoupled in several phases. First, *Network Preparation* where *FINN* uses different transformations on the *ONNX* graph to simplify it or make it more convenient for the next steps. Then, *IP Generation* where the Vivado tool is called to generate a network of High-Level Synthesis (HLS) layers with one Semiconductor Intellectual Property (IP) block per layer and finally stitches all the blocks together. Finally, the network is deployed on an FPGA with a PYNQ shell which is an FPGA utility providing a Python environment on an FPGA. This deployment is made possible by creating a project and driver then transferring it on the board along with the bitfile.

### B. Quantization

The idea of quantization is to reduce 32- or 64-bits continuous values to discrete values using a reduced amount of bits. In a neural network, the two types of values that can be quantized are the weights and the activations. Weights are the "shifting" objects in neural networks. While their initial value is either null or randomized, it is then modified by each backpropagation pass of the training phase. Activations on the other side consist of non-linear function outputs that allows simple networks to perform better on non-linear data.

The quantization of weights is done using the provided bit-width to determine a minimum and maximum integer values to perform clamping and scaling to redistribute the floating-point values to their nearest rounded $k$-bit representation. On the other hand, the quantization of activations is done using successive thresholding. For each pair of possible activation, a thresholding value is used in between. When a new value has to be run through the activation, it is simply compared to the thresholds and returns the number of threshold value that the input exceeds. The initial value in the thresholds comes from how the quantized activation is trained in *Brevitas*. *FINN* then performs a series of transformation on the trained network representation and among these transformations is a *streamline process* [23]. This process will scale and bias thresholds using scaling factors and results from the *Batch Normalization* layers and finally will round them up to the nearest number in the given representation.

### C. Folding

There is a parallel that needs to be made between the software implementation of a neural network and its hardware counterpart. The software version consists of a succession of layers with a data instance being passed and transformed from a layer to the next. The hardware architecture consists of dataflow with streaming. HLS streams (FIFOs) are used to communicate between layers and each layer is always running, waiting to perform its computation as soon as possible. The number of PE, SIMD and the FIFO depth can be chosen by hand and set in the *FINN* workflow.

Defining specific folding factors allows the user to obtain the most out of its FPGA considering the incoming network architecture and quantization. While, in this paper, those factors have remained constant for comparison purposes, they should be tuned according to the desired performance. This sizing will soon be automated inside of *FINN*. This will be done by running a simulation of the Register Transfer Level and determining the maximal occupancy of FIFOs.

Folding and parallelization will occur for specific layers such as the *Fully-Connected* ones. The two controllable degrees of parallelism are the number of *Multiply Accumulate (MACs)* inside a single product through the *Single Instruction Multiple Data (SIMD)* and the inner products in parallel through the *Physical Element (PE)*. The software representation of this parallelism can be seen on Figure 3 where the two levels can be used to process the pixels of an image in parallel in an *FC* layer. Its direct counterpart is presented on Figure 4 where the different SIMD lanes and PEs translate the parallelism.

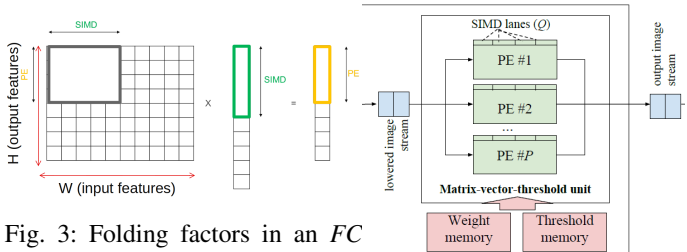

Fig. 3: Folding factors in an *FC* layer [24].

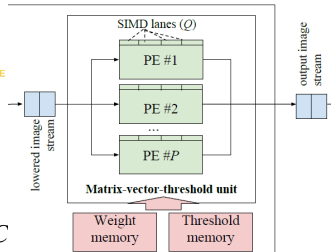

Fig. 4: Folding in Hardware [24].

## III. EXPERIMENTS

### A. Benchmark Implementation and hardware target

In order to design a benchmark of the framework and to assess the impact of quantization, there is a need for another layer on top of *Brevitas* that should be able to train, evaluate and export neural networks. While Xilinx developers provide several examples to train and export neural networks, they do not respect a common API nor the same structure rules. They are bundled in separate projects with instructions on how to run the training again but no insight is provided on what hard-coded values mean on several quantized layers.

The different objectives of the benchmark are to **Present** a common API to train neural networks, quantized or not; be able to **Resume** the training at any point; **Log** the different parts of the training, **Define** well-known neural networks and their quantized counterparts; **Set** easy access to well-known datasets; **Provide** a common API to evaluate neural networks; be able to **Export** the trained neural network to *ONNX*.

The hardware platform used in this work is the Pynq-Z1 board including a Zynq Z-7020 device: it includes 13,300 slices, 106,400 flip-flops and 630 KB Block RAMs. The development environment uses Vivado 2019.2, *ONNX* 1.5.0, *Brevitas* and *FINN* as previously described.

### B. Experiments Methodology

*1) Topology and Dataset:* The machine learning image recognition field features a wide variety of benchmarks consisting of famous datasets against which specific network architectures are run. Most famous ones are *MNIST*, *CIFAR-10* and *Imagenet*. The number of networks and their variations is thriving as well, each year presenting new networks, outperforming the predecessors. Striving for simplicity, the choice of the network architecture is conducted towards a variation of a Multilayer Perceptron (MLP) network with three Fully-Connected (FC) layers and HardTanh activations. The dataset on which this network performs is *Fashion-MNIST* [25] dataset with images resized to $32 \times 32$ pixels. This dataset is an extension of *MNIST* [26] and makes it harder for simpler architecture to classify instances while keeping the same datatype. The network is *Multi-Layer Perceptron* using *Dropout*, *Fully-Connected (FC)* and *Batch Normalization* layers. The activation used is *Hard Hyperbolic Tangent* and is used before the *Dropout* layers.

*2) Parameters and Hyper-parameters:* Training of the different networks is conducted with the parameters and hyper-parameters shown in *Table* I.

| Name | Chosen Value | Motivation |
|---|---|---|
| Network Architecture | Multilayer Perceptron (MLP) | Simplest network architecture |
| Dataset | MNIST [26] Fashion-MNIST [25] | Simplest dataset Harder iteration |
| Epochs | 100 | [22], [27] |
| Momentum | 0.9 | [22], [27] |
| Learning Rate (LR) | 0.01 | [22], [27] |
| Scheduler | Multi-step $LR \times 0.1$ at epochs 90 and 95 | [27] |
| Batch Size | 100 instances per batch | Common choice for MLP/MNIST |
| Loss Function | Cross Entropy | Common choice for MLP/MNIST |
| Optimizer | ADAM | [28] |

TABLE I: Parameters and Hyper-parameters tuning.

Along with those parameters, the bit-widths of weights and activations are chosen from 2 to 8 bits. Inputs layers are always quantized to 8 bits: using higher inputs precisions over the other quantizations will improve the end accuracy at very little cost. The deployment step is performed through the different *FINN* transformations. The training is performed with differentiated bit-widths for weights and activations, however only the models with no differences between the weights and activations are conducted through the deployment step. While Bacchus et al. [22] differentiate the bit-widths in the deployment step, this functionality is only very recently supported in *FINN*.

Out of the 7 networks (2- to 8-bit activations and weights), only the 8-bit version was not deployed due to its size. They were all run through the *FINN* transformations before getting deployed. It is important to note that the exact same folding technique has been applied to all the different networks for comparison purposes. The same comparison could be conducted using the soon-available automated folding feature from *FINN*.

*C. Results*

*1) Accuracy over Time:* Using the configuration presented above, all the networks are trained. Every each holds a different combination of bit-widths for weights and activations. Those bit-widths range from 2 to 8 and the network using 2-bit activations and 5-bit weights will be called A2W5 for the rest of the paper. Figures 5 and 6 presents the evolution of the error rate of each network against the number of epochs used in the training. The bit-widths of activations and weights are differentiated in order to assess their impact separately.

of the 49 different implementations, a tendency emerges for the *weight* bit-width to dictate the performance while the *activation* bit-width presents no consistent variations. Figures 7 and 8 presents separately the variations in performance for several weights and activations bit-widths.

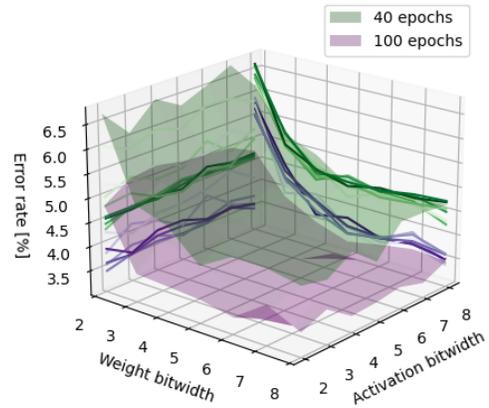

Fig. 7: Weights and activations impact for MNIST dataset.

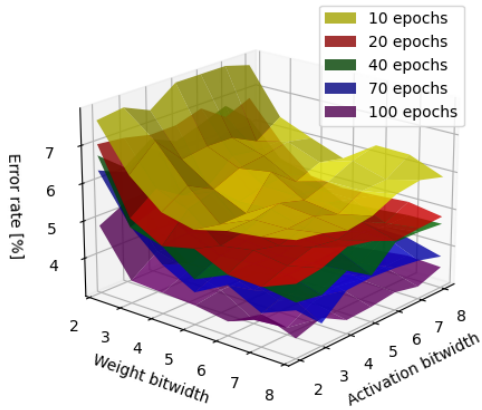

Fig. 5: Networks error rate over time for MNIST dataset.

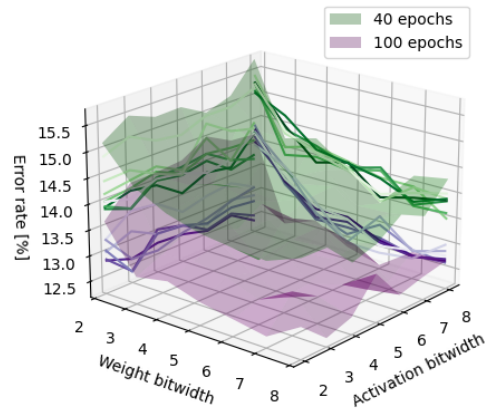

Fig. 8: Weights and activations impact for Fashion-MNIST dataset.

For clarity purposes, only two sets of networks are presented, after 40 epochs of training and after 100 epochs of training. They highlight two important results. First, the *weight* bit-width seems to be much more impactful on the overall performance than the *activation* bit-width. Next, the training time can make lower-bits implementations meet the performance of higher-bits implementations. In the example, the 64 implementations are presented after 40 and 100 epochs. The 2-bit representation obtains the same error rate after 100 epochs than the 8-bit representation after 40 epochs.

*2) Hardware Resource Utilization:* Once trained, networks with the same bit-widths for *activations* and *weights* are deployed to the PYNQ board. This deployment is done using the same folding factors for comparison purposes and results in a synthesis and implementation for each network. While the implementation is not possible for the A8W8 network, its synthesis report still allows us to look at the potential hardware utilization. The two main areas that an application uses on an FPGA are classified under logic and memory.

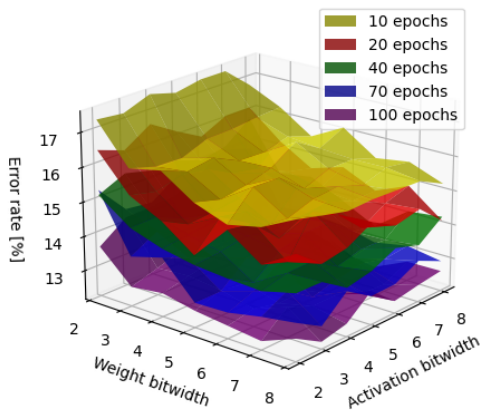

Fig. 6: Networks error rate over time for Fashion-MNIST dataset.

The more training the network gets, the more precise it becomes. The difference becomes less and less important over time and finally reaches 88%, a state-of-the-art result for this type of network. Out

Regarding logic, the number of LUTs and FFs is compared over all the implementations. The comparison is made in Figure 9. The hardware utilization on the logic side is intuitive: the higher the bit-width, the higher the number of LUTs and FFs required. The growth in hardware requirements is linear up until 6 bits where higher representations require even more resources.

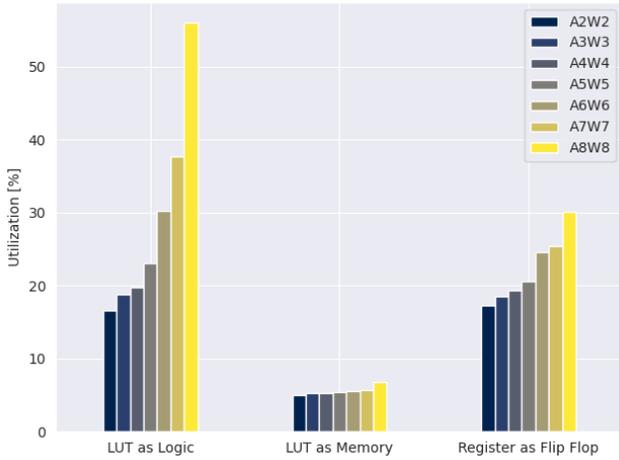

Fig. 9: Logic utilization.

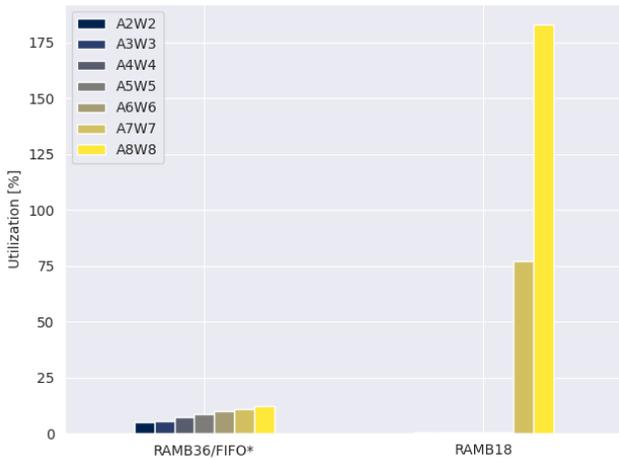

Fig. 10: Memory Utilization.

Regarding memory requirements, the amount of BRAM used is compared in Figure 10. The memory utilization shows an important use of RAMB18 when the representation exceeds 7 bits. This is due to the way *FINN* represents activations: as presented in *Section* II-B, each pair of potential output from the quantized activation is separated by a threshold value. This means that for k-bit activations, *FINN* needs $2^k-1$ threshold values and therefore that the storage for threshold values grows exponentially when the activation bit-width grows. The thresholding method is more detailed by Umuroglu et al. [23] along with the streamlining process to reduce layer operations.

*3) Throughput:* *FINN* provides a default API to perform a remote throughput test in the network once the network is deployed. This test is run with a batch size of 10,000 instances in order to measure both the throughput and DRAM In bandwidth. The ability of a network to quickly process an input directly comes from the parallelization factors set in each of the layers. As presented earlier in *Section* II-C, PE and SIMD can be chosen to determine the "folding" of each layer. In *Table* II, configurations A2W2, A3W3 and A4W4 are used with numbers of PE and SIMD both set to either 2, 8 and 16.

Note that out of all the implementations, A4W4 using parallelization factors of 16/16 could not be deployed on the target board as the design was too large to fit in the Pynq-Z1, this is represented as *Not Applicable (N/A)*. Among the different configurations (A2W2, A3W3, A4W4), no significant difference can be highlighted. The variation of PE/SIMD is the only reason for throughput variation, weights and activation bit-widths do not have a significant impact on those results. This is mainly due to the fact that the exact same folding configuration is used and that the network is small enough to be nearly identical in its 2, 3 or 4-bits variants.

| Configuration | PE/SIMD | Throughput [img/s] | DRAM In Bandwidth [Mb/s] |
|---|---|---|---|
| **A2W2** | 2/2 | 6,100 | 6.24 |
|  | 8/8 | 96,5334 | 98.85 |
|  | 16/16 | 373,112 | 382.06 |
| **A3W3** | 2/2 | 6,099 | 6.24 |
|  | 8/8 | 96,533 | 98.85 |
|  | 16/16 | 372,877 | 381.82 |
| **A4W4** | 2/2 | 6,099 | 6.24 |
|  | 8/8 | 96,512 | 98.82 |
|  | 16/16 | N/A | N/A |

TABLE II: Throughput and DRAM In bandwidths with varying PE/SIMD.

To assess the impact of PE and SIMD variations, we have taken the A3W3 configuration and created 9 variations with different combinations of PE and SIMD instances. The results can be seen in *Table* III and highlight an important fact: no clear separation can be made between the impact of the PE and SIMD values (e.g. the results of the combinations 2/8 and 8/2 or 2/16 and 16/2 are extremely similar).

| SIMD \ PE | 2 | 8 | 16 |
|---|---|---|---|
| 2 | 6,098 | 24,343 | 48,545 |
| 8 | 24,347 | 96,533 | 190,869 |
| 16 | 48,569 | 190,935 | 372,877 |

TABLE III: Throughput (*img/s*) with varying PE/SIMD in the A3W3 configuration.

The throughput logic is tightly linked to the parallelization factors since the higher they are, the higher the throughput and DRAM bandwidth will be. A3W3 still comes out as a good compromise since it allows parallelization factors up to 16 for both PE and SIMD simultaneously and therefore gets the highest output possible with a correct accuracy as stated earlier.

*D. Discussion*

These experiments confirm observations made by Bacchus et al. [22] that no important improvement is made in accuracy using beyond 3-bit representations. The eventual difference in accuracy can be covered up with enough training. The experiments highlight that:

- Accuracy is linked to the weight bit-width: the higher the weight precision, the higher the overall accuracy.
- Activation bit-width has less importance on the overall accuracy.
- Hardware utilization grows exponentially due to the *successive thresholding* method used for activation quantization.
- Training can cover up accuracy gaps between more and less precise representations. For example, the 2-bit weights implementations trained on *Fashion-MNIST* reach the same accuracy after 100 epochs than their 8-bit counterparts after 40 epochs.

- Throughput and DRAM bandwidth are tied to the parallelization factors, the higher the PE and SIMD, the higher the throughput and bandwidth.

An important note is that the *Folding Factors* used have remained the same for all the deployed implementations. While this was done for comparison purposes, it might not show the best version of lower bit-width representations. Those representations often use a small part of the target board and most of them (especially the 2- and 3-bit representations) could be better parallelized. Overall, the lowest representation implementations outperform most of their 7- or 8-bit counterparts in terms of hardware utilization, throughput and can reach a tie in accuracy given enough training.

## IV. Conclusion

CNNs are getting more and more complex in terms of architecture and costly in terms of training, inference and deployment. Interest on compression techniques has grown up in the past years. Porting ever-growing network architectures on low-power low-memory hardware architectures is still of key interest. While several methods to both and separately quantize and deploy neural networks on specific hardware architectures, very few benchmarks have been led to compare the different quantization and deployment benchmarks among them [20], [21].

This paper presents an open-source (github.com/QDucasse/nn_benchmark) and extensible trainer, built over *FINN* [16], [17] and *Brevitas*, to train QNNs on any dataset. This tool is used to assess the impact of quantization on 49 implementations of an MLP network trained on both *MNIST* and *Fashion-MNIST*. While only a subset of those implementations has been deployed on an FPGA board, results still show that enough training on a lower-precision network will make it comparable to its 8-bits counterpart while keeping hardware utilization low and throughput high. **A3W3** seems to be the best compromise between the different representations.

Future works may come from several steps out of the experiments. First, using well-known network architectures is extremely important as they have been thoroughly studied and benchmarked. The same goes for the datasets and requires no effort from the *Brevitas* development team as it is already available in *PyTorch*. While any network can be trained on any dataset for now, the issue comes from the latter parts of the workflow. *FINN* does not cover all the layer configurations for now and additional benchmarks will need to be conducted to cover the new architectures and their coupled datasets. On the other hand, comparison with classic quantization and compression frameworks (i.e. that do not require to be deployed on an FPGA board) would be interesting. Working with Intel Distiller [7] for example or the newly added quantization packages in *PyTorch* or *TensorFlow* could help broaden the comparison.

## V. Acknowledgments

This project has been supported by the French Directorate General of Armaments (DGA), the European Regional Development Fund (ERDF) of the European Union, the Brittany Region (France), the Departmental Council of Finistere and Brest Metropole as part of the Cyber-SSI project within the framework of the Brittany 2015-2020 State-Region Contract (CPER).